\documentclass[journal]{IEEEtran}
\usepackage{latexsym}

\usepackage{enumerate}
\usepackage{graphicx}
\usepackage{subfigure}
\usepackage{multirow}

\usepackage{framed,multirow}
\usepackage{times}
\usepackage{graphicx} % more modern
\usepackage{subfigure}
\usepackage{amssymb}
\usepackage{latexsym}

\usepackage{algorithm}
\usepackage{algorithmic}

\usepackage{hyperref}

\usepackage{url}
\usepackage{xcolor}
\usepackage{amsmath}
\usepackage{amsxtra}
\usepackage{braket}
\usepackage{array}
\usepackage{color}
\usepackage{url}
\usepackage{placeins}

\usepackage{times}
\usepackage{balance}
\usepackage{enumerate}
\usepackage{multirow}
\usepackage{pifont}
\usepackage{marvosym}
\usepackage{ifsym}
\usepackage{threeparttable}

\usepackage{lipsum} % 输出随机文本
\usepackage{amsmath,esint} % 输出数学公式
\usepackage{cuted}

\begin{document}
\title{AKBR: Learning Adaptive Kernel-based Representations for Graph Classification}
\author{Feifei~Qian,~\IEEEmembership{}
        Lixin~Cui,~\IEEEmembership{IEEE~Member}
        Ming~Li,~\IEEEmembership{IEEE~Member}
        Yue~Wang,~\IEEEmembership{}
        Hangyuan~Du,~\IEEEmembership{}
        Lixiang~Xu,~\IEEEmembership{}\\
        Lu~Bai${}^{*}$,~\IEEEmembership{IEEE~Member}
        Philip S.~Yu,~\IEEEmembership{IEEE~Fellow}
        Edwin R.~Hancock,~\IEEEmembership{IEEE~Fellow}

\thanks{

Feifei Qian and Lu Bai (${}^{*}$Corresponding Author: bailu@bnu.edu.cn) are with School of Artificial Intelligence, Beijing Normal University, Beijing, China. Lixin Cui and Yue Wang are with School of Information, Central University of Finance and Economics, Beijing, China. Ming Li is with the Key Laboratory of Intelligent Education Technology and Application of Zhejiang Province, Zhejiang Normal University, Jinhua, China. Hangyuan Du is with School of Computer and Information Technology, Shanxi University, Taiyuan, China.
Philip S. Yu is with Department of Computer Science, University of Illinois at Chicago, US.Edwin R. Hancock is with Department of Computer Science, University of York, York, UK}% <-this % stops a space
}
\markboth{}%
{Shell \MakeLowercase{\textit{et al.}}: Bare Demo of IEEEtran.cls for Journals}

\maketitle

\begin{abstract}
In this paper, we propose a new model to learn Adaptive Kernel-based Representations (AKBR) for graph classification. Unlike state-of-the-art R-convolution graph kernels that are defined by merely counting any pair of isomorphic substructures between graphs and cannot provide an end-to-end learning mechanism for the classifier, the proposed AKBR approach aims to define an end-to-end representation learning model to construct an adaptive kernel matrix for graphs. To this end, we commence by leveraging a novel feature-channel attention mechanism to capture the interdependencies between different substructure invariants of original graphs. The proposed AKBR model can thus effectively identify the structural importance of different substructures, and compute the R-convolution kernel between pairwise graphs associated with the more significant substructures specified by their structural attentions. Since each row of the resulting kernel matrix can be theoretically seen as the embedding vector of a sample graph, the proposed AKBR model is able to directly employ the resulting kernel matrix as the graph feature matrix and input it into the classifier for classification (i.e., the SoftMax layer), naturally providing an end-to-end learning architecture between the kernel computation as well as the classifier. Experimental results show that the proposed AKBR model outperforms existing state-of-the-art graph kernels and deep learning methods on standard graph benchmarks.
\end{abstract}


% make the title area
\maketitle
\IEEEpeerreviewmaketitle

\section{Introduction}
Graph-based representations are powerful tools for encapsulating structured data characterized by pairwise relationships among its components~\cite{zambon2018concept}, and have been widely employed in various research fields, such as the analysis of social networks \cite{bai2020learning}, financial transactions \cite{bai2020entropic} and biological networks \cite{gasteiger2021gemnet}. The main challenge arising in graph data analysis is how to learn representative numeric characteristics for discrete graph structures. One of the most effective methods for learning graph-structured data is to employ graph kernels.

Broadly speaking, graph kernels aim to describe the structural information in a high-dimension Hilbert space, typically defining a positive definite similarity measure between graphs. In 1999, Haussler~\cite{haussler1999convolution} proposed a generic way, namely the R-convolution framework, to define graph kernels. This is achieved by decomposing two graphs into substructures and evaluating the similarity between the pairs of substructures. Specifically, given two sample graphs $G_p$ and $G_q$, assume $\mathbf{S}=\{g_{1},...,g_M\}$ is the set of their all possible substructures based on a specified graph decomposing approach, the R-convolution kernel $K_R$ between $G_p$ and $G_q$ is defined as
\begin{equation}
K_R(G_p, G_q) = \sum_{\vec{g_p}\in \mathbf{S}} \sum_{\vec{g_q}\in \mathbf{S}} k(\vec{g_p},\vec{g_q}),\label{eq1}
\end{equation}
where the function $k$ is defined as the Dirac kernel, and $k(\vec{g_p},\vec{g_q})$ is equal to $1$ if the substructures $\vec{g_q}$ and $\vec{g_q}$ are isomorphic to each other, and $0$ otherwise.

In recent years, the R-convolution framework has proven to be an effective way to define novel graph kernels and most state-of-the-art R-convolution graph kernels can be categorized into three main categories, i.e., the R-convolution kernels based on the walks, paths, and subgraph or subtree structures. For instance, Gartner et al.,~\cite{gartner2003graph} have proposed a Random Walk Graph Kernel (RWGK) based on the similarity measures between random walks. Since the RWGK kernel relies on identifying isomorphic random walks between each pair of graphs associated with their directed product graph, this kernel usually requires expensive computational complexity. Moreover, the random walks suffer from the notorious tottering problem and allow the repetitive visiting of vertices, leading to significant information redundancy for the RWGK kernel. To overcome the shortcomings of the RWGK kernel, graph kernels based on paths have been developed. For instance, Borgwardt et al.,~\cite{borgwardt2005shortest} have proposed a Shortest Path Graph Kernel (SPGK) by counting the pairs of shortest paths with the same length. Since the shortest paths are typically non-backtrack paths and can be computed in a polynomial time, the SPGK kernel can significantly overcome the drawbacks of the RWGK kernel. Unfortunately, both the shortest path and the random walk are structurally simple, the resulting SPGK and RWGK kernels can only reflect limited structure information.

To overcome the above problem, more complicated substructures need to be adopted to capture more structural information, thus some subgraph-based or subtree-based R-convolution graph kernels have been developed. For instance, Shervashidze et al.,~\cite{shervashidze2009efficient} have proposed a Graphlet Count Graph Kernel (GCGK) by counting the frequency of graphlet subgraphs of sizes 3, 4 and 5. Since the GCGK kernel cannot accommodate the vertex attributes, Shervashidze et al.,~\cite{shervashidze2011weisfeiler} have further developed the Weisfeiler-Lehman Subtree Kernel (WLSK) based on subtree invariants. Specifically, the WLSK kernel first assigns an initial label to each vertex, then each vertex label is updated by mapping the sorted sets of its neighboring vertex labels into the new label. These procedures are repeated until the condition meets to the largest iteration. Since the new labels from the different iterations correspond to the subtree invariants of different heights, the WLSK kernel is defined by counting the number of pairwise isomorphic subtrees through the new labels, naturally realizing labeled graph classification. Moreover, since the WLSK kernel can efficiently and gradually aggregate the local topological substructure information (i.e., the vertex labels corresponding to subtree invariants) between neighbor vertices to further extract subtrees of large sizes, this kernel not only has better computational efficiency but also has superior effectiveness for graph classification, being one of the most popular graph kernels by now. Other graph kernels based on the R-convolution also include: (1) Optimal Assignment Kernel~\cite{kriege2016valid}, (2) the Wasserstein Weisfeiler-Lehman Subtree Kernel~\cite{togninalli2019wasserstein}, (3) the Subgraph Alignment Kernel~\cite{kriege2012subgraph}, etc.

Although state-of-the-art R-convolution graph kernels have demonstrated their performance on graph classification tasks, they still suffer from three common problems. First, these R-convolution graph kernels only focus on measuring the similarity or the isomorphism between all pairs of substructures, completely disregarding the importance of different substructures. As a result, some redundant structural information that is unsuitable for graph classification may also be considered. Second, these R-convolution graph kernels focus solely on the similarity between each pair of graphs, neglecting the common patterns shared among all sample graphs. Third, all these R-convolution graph kernels tend to employ the C-SVM classifier~\cite{cortes1995support} for classification, and the phase of training the classifier is entirely separated from that of the kernel construction, i.e., it cannot provide an end-to-end graph kernel learning framework. This certainly influences the classification performance of existing R-convolution graph kernels. To overcome the first shortcoming, Aziz et al.,~\cite{aziz2020feature} have employed the feature selection method to discard redundant substructure patterns associated with zero for the GCGK kernel, significantly improving the classification performance. However, this kernel method requires manually enumerating all possible graphlet substructure sets to compute the mean and variance, and still cannot provide an end-to-end learning framework to adaptively compute the kernel-based similarity. Overall, defining effective kernel-based approaches for graph classification remains challenging.

The objective of this work is to address the drawbacks of the aforementioned R-convolution graph kernels, by developing a novel framework to compute the Adaptive Kernel-based Representations (AKBR) for graph classification. One key innovation of the proposed AKBR model is that it can provide an end-to-end kernel-based learning framework to discriminate significant substructures and thus compute an adaptive kernel matrix between graphs. The main contributions are summarized as threefold.

\textbf{First}, to resolve the problem of ignoring the importance of different substructures that arise in existing R-convolution graph kernels, we propose to employ the attention mechanism as a means of feature selection to assign different weights to the substructures represented as features. In other words, we model the interdependency of different substructure-based features in the feature-channel attention mechanism to focus on the most essential part of the substructure-based feature vectors of graphs.

% Compared to DGCNN~\cite{zhang2018end}, we consider all local feature invariants instead of discarding some features directly.

\textbf{Second}, with the above substructure attention mechanism to hand, we define a novel kernel-based learning framework to compute the Adaptive Kernel-based Representations (AKBR) for graphs. This is achieved by computing the R-convolution kernel between pairwise graphs associated with the discriminative substructure invariants or features identified by the attention mechanism. Inspired by the graph dissimilarity or similarity embedding method presented by Bunke and Bai et al.,~\cite{bunke2008graph,bai2013graph}, the resulting kernel matrix
can be seen as a kind of kernel-based similarity embedding vectors of all sample graphs, with each row of the kernel matrix corresponds to the embedding vector of a corresponding graph. Thus, the kernel matrix can be directly input into the classifier for classification (i.e., the SoftMax layer), naturally providing an end-to-end learning
architecture over the whole procedure from the initial substructure attention layer to the final classifier. As a result, the proposed AKBR model can adaptively discriminate the structural importance of different substructures, and further compute the adaptive kernel-based representations for graph classification, significantly overcoming the three aforementioned theoretical drawbacks arising in existing R-convolution graph kernels.

\textbf{Third}, we evaluate the proposed AKBR model on graph classification tasks. The experimental results demonstrate that the proposed model can significantly outperform state-of-the-art graph kernels and graph deep learning methods.

% This paper is organized as follows. Section 2 reviews classical methods related to this work and analyze the drawbacks. Section 3 gives the definition of the AKBR model. Section 4 gives the experiments. Section 5 concludes this paper.

\section{Related Works}\label{s2}

In this section, we review some state-of-the-art R-convolution kernels that are related to our work. Moreover, we review some classical Graph Neural Networks (GNNs). Finally, we theoretically analyze the drawbacks arising in these existing approaches, enlightening the proposed method.

\subsection{Classical R-convolution Kernels}
We briefly review two classical R-convolution kernels, including the Weisfeiler-Lehman Subtree Kernel (WLSK)~\cite{shervashidze2011weisfeiler} and the Shortest Path Graph Kernel (SPGK)~\cite{borgwardt2005shortest}. We commence by introducing the definition of the WLSK kernel that focuses on aggregating the structural information from neighboring vertices iteratively to capture subtree invariants through the classical Weisfeiler-Lehman Subtree-Invariant (WL-SI) method~\cite{UWL}. Given two sample graphs $G_p$ and $G_q$, assume $l_0(u)$ represents the initial label of vertex $u$. Specifically, for unlabeled graphs, the degree of each vertex is considered as the initial label. Then, for each iteration $i$, the WLSK constructs the multi-set label $\mathcal{L}^{i}_\mathcal{N}$ for each vertex $u$ by aggregating and sorting the labels of $u$ as well as its neighborhood vertices, i.e.,
\begin{equation}
    \mathcal{L}^{i}_\mathcal{N}(u) = \mathrm{sort}(\{l_{i-1}(v)|  v\in \mathcal{N}(u)\}),
\end{equation}
where $\mathcal{N}(u)$ is the set of the neighborhood vertices of $u$. The WLSK kernel merges the multi-set label $\mathcal{L}^{i}_\mathcal{N}$ of each vertex $u$ and into a new label $l_i(u)$ through a Hash function as
\begin{equation}
    l_i(u) = \mathrm{Hash}(l_{i-1}(u), \mathcal{L}^{i-1}_\mathcal{N}(u)),\label{WLTI}
\end{equation}
where $\mathrm{Hash}$ is the hash mapping function that relabels $\mathcal{L}^{i}_\mathcal{N}$ as a new single positive integer, and each $l_i(u)$ corresponds to a subtree rooted at $u$ of height $i$. The iteration $i$ ends when the number of iterations is met to the largest one (i.e., $I_\mathrm{max}$). The WLSK kernel $K_{\mathrm{WL}}(G_p, G_q)$ between the pair of graphs $G_p$ and $G_q$ can be defined by counting the number of shared pairwise isomorphic subtrees corresponding by $l_i(u)$, i.e.,
\begin{equation}
K_{\mathrm{WL}}(G_p, G_q) = \sum_{i=0}^{I_{\mathrm{max}}} \sum_{j=0}^{|\mathcal{L}^i|}\mathcal{N}(G_p, l_i^j)\mathcal{N}(G_q, l_i^j), \label{eq01}
\end{equation}
where $I_{\mathrm{max}}$ denotes the maximum number of the iteration $i$, $l_i^j\in \mathcal{L}^i$ is the $j$-th vertex label of $\mathcal{L}^i$, and $\mathcal{N}(G_p, l_i^j)$ represents the number of the subtrees corresponded by the label $l_i^j$ and appearing in $G_p$.

The idea of the SPGK kernel is to compare the similarity between a pair of graphs by counting the number of shared shortest paths with the same lengths. The first step of computing the SPGK kernel is to extract all shortest paths from each graph by using the classical Floyd algorithm~\cite{floyd1962algorithm}. Given the pair of graphs $G_p$ and $G_q$, the SPGK is defined as
\begin{equation}
K_{\mathrm{SP}}(G_p, G_q) = \sum_{s_i\in \mathcal{S}} \mathcal{N}(G_p, s_i) \mathcal{N}(G_q, s_i), \label{eq02}
\end{equation}
where $s_i\in \mathcal{S}$ is the shortest path of length $i$, $\mathcal{S}$ is the set of all possible shortest paths appearing in all graphs, and $\mathcal{N}(G_p, s_i)$ represents the number of $s_i$ appearing in $G_p$.

\textbf{Remarks:} Although the WLSK and SPGK kernels associated with the C-Support Vector Machine (C-SVM)~\cite{ChangLinSVM2001} have effective performance for graph classification, they still have some serious theoretical drawbacks that also arise in other classical R-convolution graph kernels. First, Eq.(\ref{eq01}) and Eq.(\ref{eq02}) indicate that both the WLSK and the SPGK kernels focus on all pairs of isomorphic substructures, without considering the importance of different substructures. Since some substructures may be redundant and ineffective to discriminate the structural information between graphs, this drawback will significantly influence the classification performance. Second, the WLSK and SPGK kernels only consider the similarity measure between each individual pair of graphs, ignoring the common patterns shared among all sample graphs in the dataset.
Third, since the computation of the kernel matrix is separated from the training process of the C-SVM classifier, the kernel matrix can not be changed once the substructure invariants have been extracted. As a result, both the WLSK and SPGK kernels cannot provide an end-to-end learning architecture to adaptively compute the kernel matrix, limiting the effectiveness of existing R-convolution kernels. % In this paper, we aim to propose an Adaptive Kernel-based Representation (AKBR) model to overcome these problems.

\begin{figure*}
\centering
\includegraphics[width=0.89\linewidth]{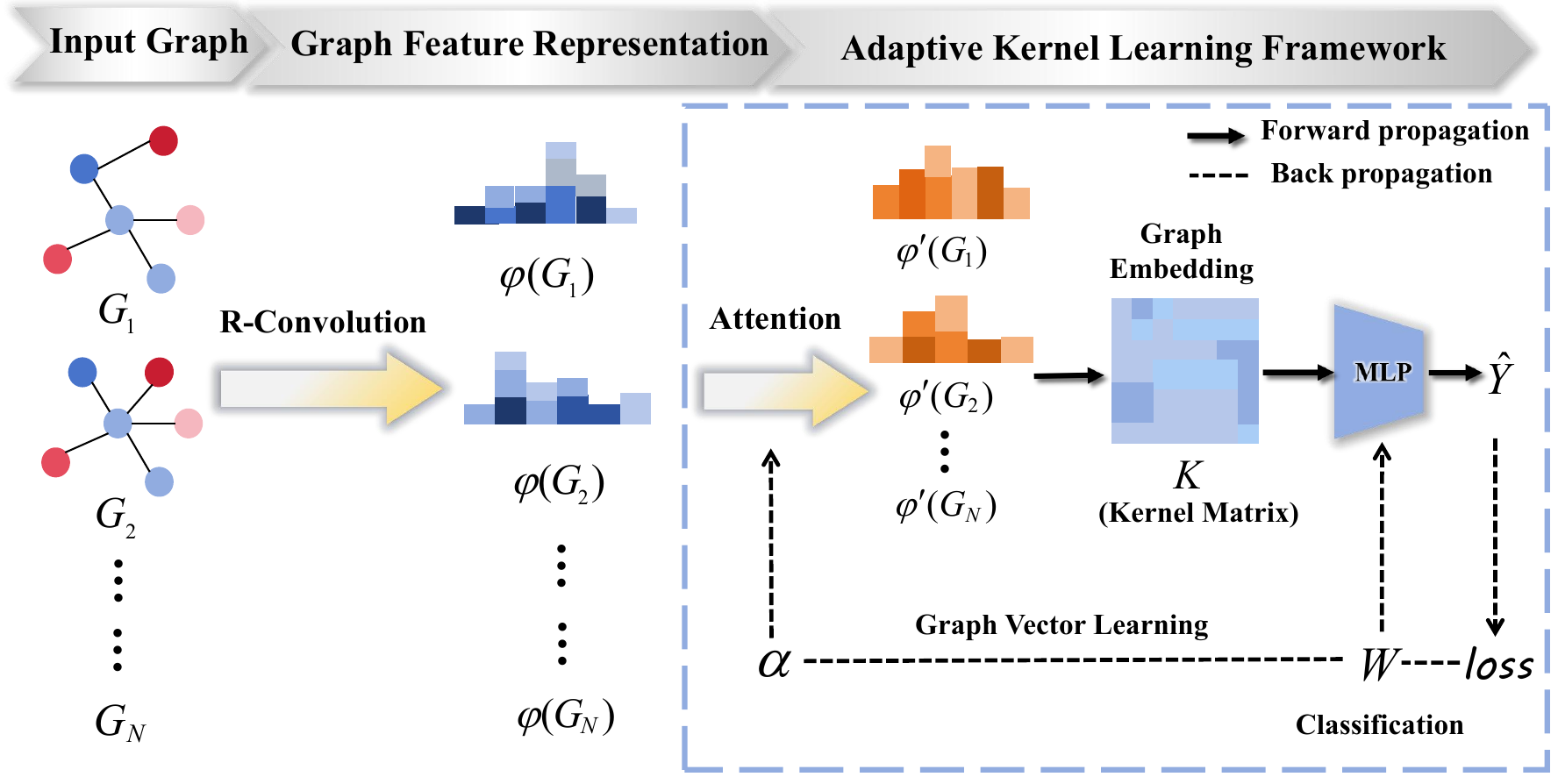}
% \vspace{-10pt}
\caption{The Framework of the Proposed AKBR Model.} \label{fig1}
% \vspace{-10pt}
\end{figure*}

\subsection{Classical Graph Neural Networks}\label{s2:2}

One way to overcome the aforementioned problems of R-convolution graph kernels is to adopt the GNN models, that are developed by generalizing the classical Deep Neural Networks (DNNs) to the graph domain based on the spectral or spatial strategy. Since the spectral-based GNN models usually require that the graphs should have the same size, and are only suitable for node classification~\cite{bruna2013spectral}. The spatial-based GNN models are widely developed for graph classification. Under this scenario, Xu et al.~\cite{DBLP:conf/iclr/XuHLJ19} have developed the Graph Isomorphism Network (GIN) that has the same expressive power with the WLSK kernel in terms of capturing the substructure information. James et al.~\cite{atwood2016diffusion} have proposed the Diffusion Convolution Neural Network (DCNN) that uses different weights to propagate the neighborhood information from different hops to the center. Niepert et al.~\cite{niepert2016learning} have developed the Graph Convolution Neural Network (PATCHY-SAN) to extract the features by converting the graph structure into fixed-sized patches, so that the standard convolution operation can be directly employed. Zhang et al~\cite{zhang2018end} have proposed the Deep Graph Convolution Neural Network (DGCNN) that directly propagates the node information through the adjacency matrix. Moreover, the DGCNN model sorts the nodes based on the substructure information extracted from the last graph convolutional layer and preserves predetermined numbers of nodes, resulting in the fixed-sized grid structure for the traditional convolution operation.

\textbf{Remarks:} Although the above GNNs can naturally provide an end-to-end framework between the substructure information extraction (i.e., the graph convolution operation) and the classifier (i.e., the SoftMax) based on the deep learning architecture that is not available for classical R-convolution kernels. These GNNs still suffer from some similar drawbacks with the R-convolution kernels. First, to provide the fix-sized graph representation for the classifier, the GIN and DCNN models tend to directly sum up the local vertex features as the global graph representation through the SumPool operation, discarding the importance of the different local structure information residing on different nodes. Second, although the PATCHY-SAN and DCNN model can form the fix-sized grid structure for the classifier, they only preserve the local structure information residing on the top ranked nodes based on the SortPool operation, resulting in significant information loss. In fact, these drawbacks also appear in other alternative GNNs, influencing the performance for graph classification.

%
%he reason is that these approaches directly use summed vertex features as the representation of global graph structures. To overcome the drawback of these models associated with SumPooling while ensuring permutation invariance, sorted-based methods are proposed
%
%
%Although these sorted-based methods have dealt with the problem of permutation invariance, information loss usually arises when transforming the topology structures into fixed-sized grids.

\section{The Proposed AKBR Model for Graphs}
In this section, we develop a novel Adaptive Kernel-based Representations (AKBR) model. We commence by introducing the detailed definition of the proposed AKBR model. Moreover, we discuss the advantages of the proposed AKBR model, explaining its effectiveness.

\subsection{The Framework of the AKBR Model}
In this subsection, we define the framework of the proposed AKBR approach. Specifically, the computational architecture of the proposed AKBR model is shown in Fig.~\ref{fig1}, mainly consisting of four procedures.

\textbf{For the first step}, we construct the feature vector $\varphi(G_i)$ for each sample graph $G_i$ based on the substructure invariants extracted with a specific R-convolution graph kernel, and each element of the feature vector corresponds to the number of a corresponding substructure appearing in the graph. In this work, we propose to adopt the classical WLSK and SPGK kernels for the framework of the proposed model. This is because the subtree and shortest path invariants associated with the two kernels can be efficiently extracted from original graph structures. Moreover, both the WLSK and SPGK kernels have effective performance for graph classification, indicating that their associated substructures are effective in representing the structural characteristics of original graphs.

\textbf{For the second step}, unlike the classical WLSK and SPGK kernels that are computed based on all possible specific substructures, we adaptively select a family of relevant substructures for the proposed AKBR model, i.e., we propose to select the most effective features for the graph feature vector $\varphi(G_i)$. This is based on the fact that some substructures are redundant or not effective in reflecting the kernel-based similarity between pairwise graphs~\cite{aziz2020feature}, influencing the performance of the R-convolution kernels. To this end, we employ an attention layer to assign different weights to the substructure features, and the critical features will be associated with larger weights through the attention mechanism, resulting in an attention-based feature vector for each graph.

\textbf{For the third step}, based on the attention-based feature vectors of all graphs computed from the second step, the resulting kernel matrix between pairwise graphs can be computed as the dot product between their attention-based substructure feature vectors.

\textbf{For the fourth step}, inspired by the graph dissimilarity or similarity embedding method presented by Bunke and Bai et al.,~\cite{bunke2008graph,bai2013graph}, we employ the resulting kernel matrix from the third step as the kernel-based similarity embedding vectors of all sample graphs, where each row of the kernel matrix corresponds to the embedding vector of a corresponding graph. We directly input the kernel matrix into the classifier for classification. To provide an end-to-end learning framework for the proposed AKBR model, we propose to employ the Multi-Layer Perceptron classifier (MLP) for classification, and the MLP consists of two fully connected layers associated with an activation function (i.e., the SoftMax).

The output of the MLP is the predicted graph label $\hat{Y}$, and the loss is the error between the predicted label $\hat{Y}$ and the real graph label $Y$ using the cross-entropy. The \textbf{\emph{attention-based weights}} for the features of the feature vector $\varphi_{(\cdot)}(G_i)$ and the \textbf{\emph{trainable parameter matrix}} for the MLP will be updated when the loss is backpropagated. As a result, the framework of the proposed AKBR model can provide an end-to-end learning architecture between the kernel computation as well as the classifier, i.e., the proposed AKBR model can adaptively compute the kernel matrix associated with the most effective substructure invariants.

\subsection{The Definition of the AKBR Model}
In this subsection, we give a detailed definition of the four computational steps described in Section~3.1. Specifically, Section 3.2.1 introduces the construction of graph feature vectors based on substructure invariants. Section 3.2.2 introduces the feature-channel attention mechanism for feature selection. Subsequently, the construction of the adaptive kernel matrix is presented in Section 3.2.3. Finally, Section 3.2.4 shows how the kernel matrix can be seen as a kind of similarity embedding vector of graph structures for classification.

\subsubsection{\textbf{The Construction of Substructure Invariants}}
\
\newline
We employ the classical WLSK and SPGK kernels to extract the subtrees and the shortest paths as the substructure invariants. For the WLSK kernel, the subtree-based feature vector $\varphi_{\mathrm{WL}}(G)$ of a sample graph $G$ is defined as
\begin{equation}
    \varphi_{\mathrm{WL}}(G) = [\mathrm{n}(G, l_1),\ldots, \mathrm{n}(G, l_i),\ldots, \mathrm{n}(G, l_{|{\mathcal{L}}|})], \label{eq00}
\end{equation}
where $l_i$ is the vertex label defined by Eq.(\ref{WLTI}) and corresponds to a subtree invariant, each element $\mathrm{n}(G, l_i)$ is the number of the corresponding subtree invariants appearing in $G$, and $|\mathcal{L}|$ is a positive integer and refers to the number of all distinct subtree invariant labels. Similarly, for the SPGK kernel, the feature vector $\varphi_{\mathrm{SP}}(G)$ of the graph $G$ is defined as
\begin{equation}
    \varphi_{\mathrm{SP}}(G) = [\mathrm{n}(G, s_1),\ldots, \mathrm{n}(G, s_i),\ldots, \mathrm{n}(G, s_{|S|})],
\end{equation}
where each element $\mathrm{n}(G, s_i)$ is the number of the shortest paths with the length $s_i$ in the graph $G$, and $|S|$ denotes the greatest length of the shortest paths over all graphs. With the substructure-based feature vectors of all graphs in $\mathbf{G}=\{G_1,\ldots, G_N\}$ to hand, we can derive the feature matrix $\mathbf{X} \in \mathbb{R}^{N \times L}$ for the entire graph dataset $\mathbf{G}$, i.e.,
\begin{equation}
    \mathbf{X}_{(\cdot)}=
    \begin{pmatrix}
    \varphi_{(\cdot)}(G_1) \\
    ... \\
    \varphi_{(\cdot)}(G_j) \\
    ... \\
    \varphi_{(\cdot)}(G_N) \\
    \end{pmatrix},\label{feature_matrix}
\end{equation}
where $N$ denotes the number of graphs in $\mathbf{G}$, $L$ denotes the dimension of each feature vector $\varphi(G_j)$ for the graph $G_j\in \mathbf{G}$, and ${(\cdot)}$ corresponds to either the WLSK or the SPGK kernel.

\begin{figure}
\centering
\includegraphics[width=0.99\linewidth]{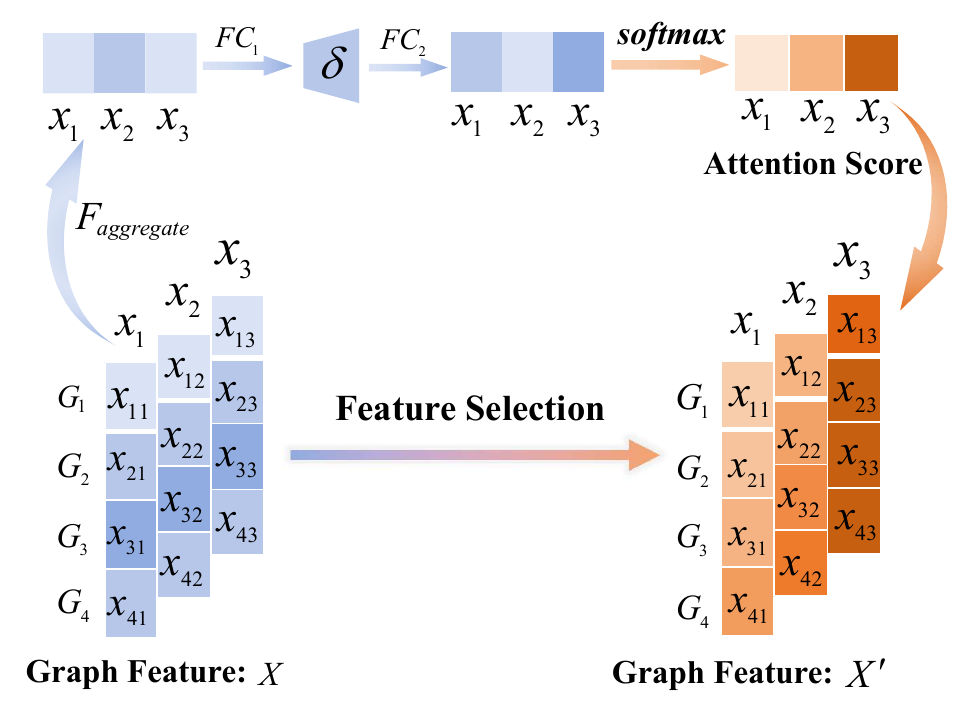}
% \vspace{-20pt}
\caption{One Illustrative Instance of the Attention Mechanism for Feature Selection.} \label{fig2}
% \vspace{-10pt}
\end{figure}
\subsubsection{\textbf{Attention Mechanism for Feature Selection}}
As we have stated previously, some substructure-based features may be more prevalent in some graphs. These features naturally encapsulate more significant and discriminative structural information for classification. Thus, assigning a more substantial weight to these features is preferred. On the other hand, some features may be less effective for classification, it is reasonable to assign smaller weights to these ineffective features. As a result, it is necessary to perform a thorough examination to evaluate the importance of different features across all graphs. To this end, we propose to employ the attention mechanism as a means of feature selection, and adaptive identify the most effective features of the feature matrix $\mathbf{X}_{(\cdot)} \in \mathbb{R}^{N \times L}$ defined by Eq.(\ref{feature_matrix}).

There have been various types of attention mechanisms, including the self-attention~\cite{vaswani2017attention}, the external attention~\cite{guo2022beyond}, and channel attention~\cite{hu2018squeeze}. Inspired by the recent attention-based work~\cite{hu2018squeeze} that proposes to squeeze the global spatial information into a channel descriptor, we commence by using the \emph{SumPooling} operation to aggregate the information of all graphs and squeeze them into a feature channel. Given the feature matrix $\mathbf{X}_{(\cdot)} \in \mathbb{R}^{N \times L}$ of all graphs in $\mathbf{G}$, the aggregation information $\mathbf{h} \in \mathbb{R}^{1 \times L}$ can be calculated as
\begin{equation}
    h_l = F_{\mathrm{aggregate}}(\boldsymbol{x_l}) = \frac{1}{N} \sum_{i=1}^{N}x_{l,i},\label{eq7}
\end{equation}
where $h_l$ is the $l$-th element of $\mathbf{h}$. With $\mathbf{h}$ to hand, we use two fully connected linear layers associated with the non-linear activation function to calculate the attention scores. Specifically, we use $W_1 \in \mathbb{R}^{L \times C}$ to denote the weight of the first fully connected layer and $W_2 \in \mathbb{R}^{C \times L}$ to represent the weight of the second dense layer, where $C$ denotes the hidden feature dimension.

The resulting attention-based scoring matrix $\mathbf{\alpha} \in \mathbb{R}^{N \times L}$ for the feature matrix $\mathbf{X}_{(\cdot)} \in \mathbb{R}^{N \times L}$ can be computed as
\begin{equation}
\mathbf{\alpha} = \mathrm{softmax}(W_2\sigma(W_1\mathbf{h})), \label{eq12}
\end{equation}
where $\sigma$ is the ReLU function, and $\mathbf{\alpha}$ records the attention scores.

With attention-based scoring matrix $\mathbf{\alpha}$ that encapsulates adaptive weights for the different features of each graph, the substructure-based feature matrix $\mathbf{X}_{(\cdot)}$ can be updated as the weighted substructure-based feature matrix $\mathbf{X'}_{(\cdot)}$ by multiplying the attention scores, i.e.,
\begin{equation}
\mathbf{X'}_{(\cdot)} = \mathbf{\alpha}  \otimes \mathbf {X}_{(\cdot)}, \label{eq13}
\end{equation}
where $\otimes$ refers to the feature-wise multiplication. An instance of the attention mechanism for feature selection is shown in Fig.\ref{fig2}. In summary, the attention mechanism can assign substructures of each graph with different weights according to their importance.
\subsubsection{\textbf{The Adaptive Construction of the Kernel Matrix}}
\
\newline
Based on the definition in~\cite{shervashidze2011weisfeiler}, any R-convolution graph kernel can be computed as the dot product between the substructure-based feature vectors of pairwise graphs. For an instance, the WLSK kernel $ K_{\mathrm{WL}}$ defined by Eq.(\ref{eq01}) between a pair of graphs $G_p$ and $G_q$ can be rewritten as
\begin{equation}
  K_{\mathrm{WL}}(G_p, G_q) = \langle \varphi_\mathrm{WL}(G_p), \varphi_\mathrm{WL}(G_q) \rangle,
\end{equation}
where each $\varphi_\mathrm{WL}(G_p)$ is the substructure-based feature vector defined by Eq.(\ref{eq00}). Thus, with the weighted substructure-based feature matrix $\mathbf{X'}_{(\cdot)}$ of all graphs in $\mathbf{G}$ defined by Eq.(\ref{eq13}) to hand, we can compute the attention-based kernel matrix $\mathbf{K}_{(\cdot)} \in \mathbb{R}^{N \times N}$ by directly dot-multiplying the feature vectors in $\mathbf{X'}$ as
\begin{equation}
\mathbf{K}_{(\cdot)} = \mathbf{X'}_{(\cdot)} \mathbf \cdot {\mathbf{X'}_{(\cdot)}}^T,\label{Attention_K}
\end{equation}
where ${(\cdot)}$ indicates that the kernel matrix $\mathbf{K}_{(\cdot)}$ can be computed using either the WLSK or the SPGK kernel.
% Since the framework of the proposed AKBR model can provide an end-to-end learning mechanism to update the attention
\subsubsection{The Kernel-based Graph Embedding Vectors}
\
\newline
We show how the attention-based kernel matrix $\mathbf{K}_{(\cdot)} $ defined by Eq.(\ref{Attention_K}) can be seen as the embedding vectors of all graphs for the classifier. Specifically, Riesen and Bunke~\cite{DBLP:series/smpai/RiesenB10} have proposed a (dis)similarity graph embedding method that can embed or convert each graph structure into a vector, so that any standard machine learning and pattern recognition for vectors can be directly employed. For a sample graph $G$ and a set of prototype graphs $\mathbf{G}^\mathrm{p}=(G^{\mathrm{p}}_1,\ldots, G^{\mathrm{p}}_m,\ldots, G^{\mathrm{p}}_M)$, the embedding vector $\phi(G)$ of $G$ is defined as
\begin{equation}
    \phi(G) = [f(G, G^{\mathrm{p}}_1),\ldots, f(G, G^{\mathrm{p}}_m),\ldots, f(G, G^{\mathrm{p}}_M)], \label{eq8}
\end{equation}
where each element $f(G, G^{\mathrm{p}}_m)$ can be a (dis)similarity measure between the sample graph $G \in \mathbf{G}$ and the $m$-th prototype graph $G^{\mathrm{p}}_m \in \mathbf{G}^\mathrm{p}$. Inspired by this graph embedding method, we propose to employ all sample graphs as the prototype graphs and use the graph kernel as the means of the similarity between each sample graph $G$ and each other graph $G_q\in \mathbf{G}$ (including $G$ itself). Thus similar to Eq.(\ref{eq8}), the kernel-based embedding vector $\phi_{(\cdot)}(G)$ of $G$ can be defined as
\begin{equation}
\phi_{(\cdot)}(G) = [K_{(\cdot)}(G, G_1),\ldots, K_{(\cdot)}(G, G_q),\ldots, K_{(\cdot)}(G, G_N) ], \label{K_embedding}
\end{equation}
where each element $K_{(\cdot)}(G, G_q)$ represents the kernel value between $G$ and $G_p$, and ${(\cdot)}$ indicates that Eq.(\ref{K_embedding}) can be computed with either the WLSK kernel or the SPGK kernel. Clearly, if $G\in \mathbf{G}$ is the $p$-th sample graph $G_p\in \mathbf{G}$ (i.e., $G = G_p$), the kernel-based embedding vector $\phi_{(\cdot)}(G)$ is essentially the $p$-th row of the kernel matrix $\mathbf{K}_{(\cdot)}$. As a result, the kernel matrix $\mathbf{K}_{(\cdot)}$ can be theoretically seen as the kernel-based embedding vectors over all graphs from $\mathbf{G}$.

We propose to directly input the kernel matrix $\mathbf{K}_{(\cdot)}$ into the MLP classifier for classification. Since the \textbf{\emph{attention-based weights}} for computing the kernel matrix $\mathbf{K}_{(\cdot)}$ and the \textbf{\emph{trainable parameter matrix}} for the MLP can be adaptively updated when the loss is backpropagated, the computational framework of the proposed AKBR model can naturally provide an end-to-end learning mechanism, that can adaptively compute the kernel matrix $\mathbf{K}_{(\cdot)}$ associated with more effective substructures for graph classification.

\subsection{Discussion of the Proposed Model}
The proposed AKBR model has several important properties that are not available for most existing R-convolution graph kernels, explaining the theoretical effectiveness.

\textbf{First}, unlike existing R-convolution graph kernels, the proposed AKBR model can assign different weights to the substructure-based features to identify the importance between different substructures, based on the attention mechanism. By contrast, the existing R-convolution graph kernels focus on measuring the isomorphism between all pairs of substructures, without considering the importance of different substructures. Thus, the proposed AKBR model can compute a more effective kernel matrix for classification.

\textbf{Second}, unlike existing R-convolution graph kernels that only reflect the similarity between each individual pair of graphs, the proposed AKBR model can capture the common patterns over all graphs in the dataset. This is because employing the attention-based feature selection for the substructure-based feature vectors needs to evaluate the effectiveness of all possible substructures over all graphs, the proposed AKBR model can potentially accommodate the structural information over all graphs. Moreover, since the proposed AKBR model is defined as associated with an end-to-end computational framework, all graphs will be used for the training, capturing the main characteristics of all graphs.

\textbf{Third}, the R-convolution graph kernels tend to employ the C-SVM classifier for classification, and the phase of training the classifier is entirely separated from that of the kernel matrix construction. As a result, these kernels cannot provide an end-to-end learning framework, and the kernel matrix can not be adaptively updated during the training process. By contrast, the proposed AKBR model is defined based on an end-to-end learning framework, the loss of the associated MLP classifier can be backpropagated to update the attention-based weights of all substructures, adaptively computing the kernel matrix for graph classification.

\textbf{Forth}, unlike the GNNs discussed in Section~\ref{s2:2}, the proposed AKBR model can either identify the importance of the different local structure information residing on all local substructure invariants or provide an end-to-end framework between the structure information extraction (i.e., the kernel-based embedding) and the classifier (i.e., the SoftMax).

% also overcome the information loss arising in existing GNNs. Moreover, similar to these GNNs, the proposed AKBR model can

\section{Experiments}
In this section, we evaluate the performance of the proposed AKBR model against state-of-the-art graph kernels and deep learning methods. We use six standard graph datasets extracted from bioinformatics (Bio), social networks (SN), and computer vision (CV). These datasets from bioinformatics and social networks can be directly downloaded from~\cite{morris2020tudataset}. The Shock dataset can be obtained from~\cite{siddiqi1999shock}. Detailed descriptions of these six datasets are shown in Table~\ref{tab1}.

\begin{table*}
\caption{Information of the graph datasets}\label{tab1}
% \vspace{-10pt}
\centering{
\footnotesize
% \scriptsize
% \tiny
\begin{tabular}{|c|c|c|c|c|c|c|}
\hline
Datasets & MUTAG  &PTC(MR) & PROTEINS & IMDB-B & IMDB-M & Shock\\
\hline \hline
Max  \# vertices & 28 & 109 & 620  & 136 & 89 & 33\\ \hline
Mean  \# vertices & 17.93& 25.56 & 39.06 & 19.77 & 13 & 13.16\\ \hline
\# graphs & 188  & 344 & 1113  & 1000 & 1500 & 150\\ \hline
\# classes& 2  & 2 & 2  & 2 & 3 &10 \\ \hline
Description & Bio  & Bio  & Bio & SN & SN & CV\\
\hline
\end{tabular}
}
% \vspace{-5pt}
\end{table*}

\begin{table*}

\caption{Classification accuracy (in \% $\pm$ standard error) comparisons with graph kernels.}\label{tab3}
% \vspace{-10pt}
\centering{
\footnotesize
% \scriptsize
\begin{tabular}{|c|c|c|c|c|c|c|}
\hline
 Datasets    & MUTAG          & PTC(MR)         & PROTEINS       & IMDB-B          & IMDB-M & Shock\\ \hline \hline
     {GCGK3} & 82.04$\pm$0.39 & 55.41$\pm$0.59  & 71.67$\pm$0.55 &  --             & --             & 26.93$\pm$0.63\\ \hline
        RWGK & 80.77$\pm$0.72 & 55.91$\pm$0.37  & 74.20$\pm$0.40 &  67.94$\pm$0.77 & 46.72$\pm$0.30 &  2.31$\pm$1.13 \\ \hline
        SPGK & 83.38$\pm$0.81 & 55.52$\pm$0.46  & 75.10$\pm$0.50 &
        71.26$\pm$1.04 & 51.33$\pm$0.57 & 37.88$\pm$0.93\\ \hline
        CORE SP & 88.29 $\pm$1.55 & 59.06 $\pm$0.93  & -- &  72.62 $\pm$0.59 & 49.43$\pm$0.42 & -- \\ \hline
       %  JTQK & 85.50 $\pm$0.55 & 58.50$\pm$0.39  & 72.86$\pm$0.41 &
       % 72.45$\pm$0.81 & 50.33 $\pm$0.49 & 37.73 $\pm$0.42 \\ \hline
        WLSK & 82.88$\pm$0.57 & 58.26$\pm$0.47  & 73.52$\pm$0.43 &  71.88$\pm$0.77 & 49.50$\pm$0.49 & 36.40$\pm$1.00\\ \hline
     CORE WL & 87.47$\pm$1.08 & 59.43$\pm$1.20  & --             &  74.02$\pm$0.42 & 51.35$\pm$0.48   & --\\ \hline
        WL-OA & 84.5$\pm$1.70 & 63.6$\pm$1.5  & 76.40$\pm$0.40 &
       -- & -- & --\\ \hline
        % ASPK & 89.42$\pm$0.08 & 55.09$\pm$0.08 & 64.02$\pm$0.07 & 38.33$\pm$0.08 & \textbf{76.64}$\pm$0.40 & 75.40$\pm$0.03 & 48.20$\pm$0.03 \\ \hline
        \textbf{AKBR (SP)} & 86.17$\pm$1.12 & \textbf{64.56$\pm$1.08} & \textbf{77.07$\pm$0.75} & \textbf{75.19$\pm$0.47} & \textbf{52.61$\pm$0.44} & \textbf{41.22$\pm$3.98}\\ \hline
\textbf{AKBR\_1(WL)} & \textbf{90.75$\pm$0.58} & \textbf{64.86$\pm$0.71}  & \textbf{77.34$\pm$0.29} &  \textbf{75.35$\pm$0.57} & \textbf{52.06$\pm$0.49} & \textbf{44.43$\pm$3.01} \\ \hline
\textbf{AKBR\_2(WL)} & \textbf{90.87$\pm$0.51} & \textbf{64.85$\pm$1.47}  & \textbf{76.38$\pm$0.48} &  \textbf{75.07$\pm$0.53}& \textbf{51.48$\pm$0.42} & \textbf{41.01$\pm$2.58} \\ \hline
\textbf{AKBR\_3(WL)} & \textbf{90.26$\pm$0.79} & \textbf{64.45$\pm$1.37}  & \textbf{76.59$\pm$0.21} &  \textbf{74.21$\pm$1.25} & \textbf{51.24$\pm$0.51} & \textbf{38.58$\pm$2.66} \\ \hline
\end{tabular}
}

\end{table*}

\begin{table*}
\caption{The Hyper-parameters for AKBR}\label{tab5}
% \vspace{-10pt}
\centering{
\footnotesize
% \scriptsize
\begin{tabular}{|c|c|c|c|c|c|c|}
\hline
Datasets   & lr     & epoch & wd            & att\_hid & nhid1 & nhid2 \\ \hline \hline
MUTAG      & 0.006  & 500   & 5.00E-08      & 50       & 150   & 300   \\ \hline
PTC(MR)    & 0.004  & 500   & 5.00E-08      & 50       & 50    & 300   \\ \hline
PROTEINS   & 0.0004 & 500   & 5.00E-06      & 50       & 50    & 300   \\ \hline
IMDB-B     & 0.006  & 500   & 5.00E-08      & 50       & 150   & 300   \\ \hline
IMDB-M     & 0.006  & 500   & 5.00E-08      & 50       & 150   & 300   \\ \hline
Shock      & 0.006  & 500   & 5.00E-08      & 50       & 50    & 300   \\ \hline
\end{tabular}}
% \vspace{-15pt}
\end{table*}

\subsection{Comparisons with Graph Kernels}

\subsubsection{\textbf{Experimental Setups}}
% We first compare three powerful R-convolution kernels with their AKBR variants as Table~\ref{tab2}:
% (1) Graphlet Count Graph Kernel (GCGK)~\cite{shervashidze2009efficient} with graphlet of size 3,
% (2)Shortest Path Graph Kernel (SPGK)~\cite{borgwardt2005shortest},
% (3) Weisfeiler-Lehman Subtree Kernel (WLSK)~\cite{shervashidze2011weisfeiler}.

We compare the performance of the proposed AKBR model with several state-of-the-art graph kernels for graph classification tasks as Table~\ref{tab3}:  (1) the Graphlet Count Graph Kernel (GCGK)~\cite{shervashidze2009efficient} with graphlet of size 3, (2) the Random Walk Graph Kernel (RWGK)~\cite{gartner2003graph}, (3) the Shortest Path Graph Kernel (SPGK)~\cite{borgwardt2005shortest} (4) the Shortest Path Kernel based on Core-Variants (CORE SP) ~\cite{nikolentzos2018degeneracy} (5) the Weisfeiler-Lehman Subtree Kernel (WLSK)~\cite{shervashidze2011weisfeiler}, and (6) the WLSK kernel associated with Core-Variants (CORE WL)~\cite{DBLP:conf/ijcai/NikolentzosMLV18}, (7) Valid Optimal Assignment Kernel (WL-OA)~\cite{kriege2016valid}. We perform a 10-fold cross-validation using the C-SVM classifier for each alternative graph kernel. We repeat the experiments ten times and the average accuracy is reported in Table~\ref{tab3}. We search the optimal hyperparameters for each graph kernel on each dataset. Since some methods are not evaluated by the original paper on some datasets, we do not provide these results. For the proposed AKBR model, we conduct the experiment based on the SPGK and WLSK kernels to demonstrate the effectiveness. We use the AKBR\_$i$(WL) to denote the AKBR model based on the WLSK kernel, with $i$ to denote the iteration parameters. Besides, we use the AKBR(SP) to represent the AKBR model based on the SPGK. The classification accuracies of the proposed AKBR model are also based on the 10-fold cross-validation strategy. Finally, we set other parameters of the proposed AKBR model for different datasets as shown in Table~\ref{tab5}, including the learning ratio ($lr$), the weighted decay ($wd$), the training epoch, hidden dimensions for the attention mechanism ($att_hid$, $nhid1$, $nhid2$). For the classifier, we employ three fully connected linear layers associated with the RELU function as the non-linear activation function.

% ReLU. The detailed parameters we used are provided in Table~\ref{tab5}.

\subsubsection{\textbf{Experimental Results and Analysis}}
Compared to the classical graph kernels, the family of the proposed AKBR models achieves highly competitive accuracies in Table~\ref{tab3}, demonstrating that the proposed AKBR framework is effective. Specifically, we observe that the family of the proposed AKBR$_i$(WL)s can outperform state-of-the-art graph kernels on all datasets. Note that, for the proposed AKBR$_i$(WL)s, we set the iteration as 1, 2, and 3, and the accuracy is still higher than the original optimal WLSK kernel. On the other hand, the proposed AKBR(SP) also significantly outperforms the original SPGK kernel. These observations demonstrate that the proposed AKBR model can adaptively identify more effective substructure-based features, and achieve better classification performance than the WLSK and the SPGK kernels. The experimental results demonstrate the theoretical advantages of the proposed AKBR model, i.e., adaptively identifying the importance of different substructures and computing the adaptive kernel matrix through an end-to-end learning framework can tremendously improve the classification performance.

\begin{table*}
\caption{Classification accuracy (in \% $\pm$ standard error) comparisons with deep learning methods.}\label{tab4}
% \vspace{-10pt}
\centering{
\footnotesize
% \scriptsize
    % \tiny
    \begin{tabular}{|c|c|c|c|c|c|}
    \hline
        Datasets & MUTAG          & PTC(MR)        & PROTEINS       & IMDB-B         & IMDB-M \\ \hline \hline
        DGCNN    & 85.83$\pm$1.66 & --             & 75.54$\pm$0.94 & 70.03$\pm$0.86 & 47.83$\pm$0.85 \\ \hline
        DCNN     & 66.98          & 56.60          & 61.29$\pm$1.60 & 49.06$\pm$1.37 & 46.72$\pm$0.30 \\ \hline
        PATCHY-SAN   & 88.95$\pm$4.37 & 62.29 $\pm$ 5.68 & 75.00$\pm$2.51 & 71.00$\pm$2.29 & 45.23$\pm$2.84 \\ \hline
        DGK      & 82.66$\pm$1.45 & 60.08 $\pm$ 2.55 & 71.68$\pm$0.50 & 66.96$\pm$0.56 & 44.55$\pm$0.52 \\ \hline

        ~1-RWNN~ & ~$89.2\pm4.3$~ & ~$-$~        & ~$70.8\pm4.8$~                   & ~$70.8\pm4.8$~ & ~$47.8\pm3.8$~  \\ \hline
        ~2-RWNN~ & ~$88.1\pm4.8$~ & ~$-$~        & ~$74.7\pm3.3$~                   & ~$70.6\pm4.4$~ & ~$48.8\pm2.9$~  \\ \hline
        ~3-RWNN~ & ~$88.6\pm4.1$~ & ~$-$~        & ~$74.1\pm2.8$~                   & ~$70.7\pm3.9$~ & ~$47.8\pm3.5$~  \\ \hline
        ~GIN~    & ~$84.7\pm6.7$~ & ~$-$~        & ~$74.3\pm3.3$~                   & ~$71.23\pm3.9$~ &~$48.53\pm3.3$~ \\ \hline

\textbf{AKBR (SP)} & 86.17$\pm$1.12 & \textbf{64.56$\pm$1.08} & \textbf{77.07$\pm$0.75} & \textbf{75.19$\pm$0.47} & \textbf{52.61$\pm$0.44} \\ \hline
\textbf{AKBR\_1(WL)} & \textbf{90.75$\pm$0.58} & \textbf{64.86$\pm$0.71}  & \textbf{77.34$\pm$0.29} &  \textbf{75.35$\pm$0.57} & \textbf{52.06$\pm$0.49}  \\ \hline
 \textbf{AKBR\_2(WL)} & \textbf{90.87$\pm$0.51} &\textbf{64.85$\pm$1.47}  & \textbf{76.38$\pm$0.48} &  \textbf{75.07$\pm$0.53} & \textbf{51.48$\pm$0.42}  \\ \hline
 \textbf{AKBR\_3(WL)} & \textbf{90.26$\pm$0.79} & \textbf{64.45$\pm$1.37}  & \textbf{76.59$\pm$0.21} &  \textbf{74.21$\pm$1.25} & \textbf{51.24$\pm$0.51}  \\ \hline
    \end{tabular}
    }
\end{table*}

\subsection{Comparisons with Deep Learning}
\subsubsection{\textbf{Experimental Settings}} We further compare the family of our proposed AKBR models with some state-of-the-art graph deep learning methods, including (1) the Deep Graph Convolution Neural Network (DGCNN)~\cite{zhang2018end}, (2) the Diffusion Convolution Neural Network (DCNN)~\cite{atwood2016diffusion}, (3) the PATCHY-SAN based Graph Convolution Neural Network (PATCHY-SAN)~\cite{niepert2016learning}, (4) the Deep Graphlet Kernel (DGK)~\cite{yanardag2015deep}, (5) the Random Walk Graph Neural Networks ($p$-RWNN)~\cite{DBLP:conf/nips/NikolentzosV20} associated with three different random walk length $p$ ($p=1,2,3$), and (6) the Graph Isomorphism Network (GIN)~\cite{DBLP:conf/iclr/XuHLJ19}. These deep learning methods were also evaluated using the same 10-fold cross-validation strategy with ours, thus we directly report the results from the original papers in Table~\ref{tab4}. Note that, Errica et al.~\cite{DBLP:conf/iclr/ErricaPBM20} have stated that some popular graph deep learning methods often lack rigorousness and are hardly reproducible. To overcome this problem, they have provided some fair experimental evaluations for these methods with the same experimental settings, and Nikolentzos et al.~\cite{DBLP:conf/nips/NikolentzosV20} also compare their $p$-RWNN model with other methods associated with the results reported in~\cite{DBLP:conf/nips/NikolentzosV20}. For a fair comparison, we also directly cite the results from~\cite{DBLP:conf/nips/NikolentzosV20} for the GIN model.

\begin{figure}
\centering
\includegraphics[width=0.99\linewidth]{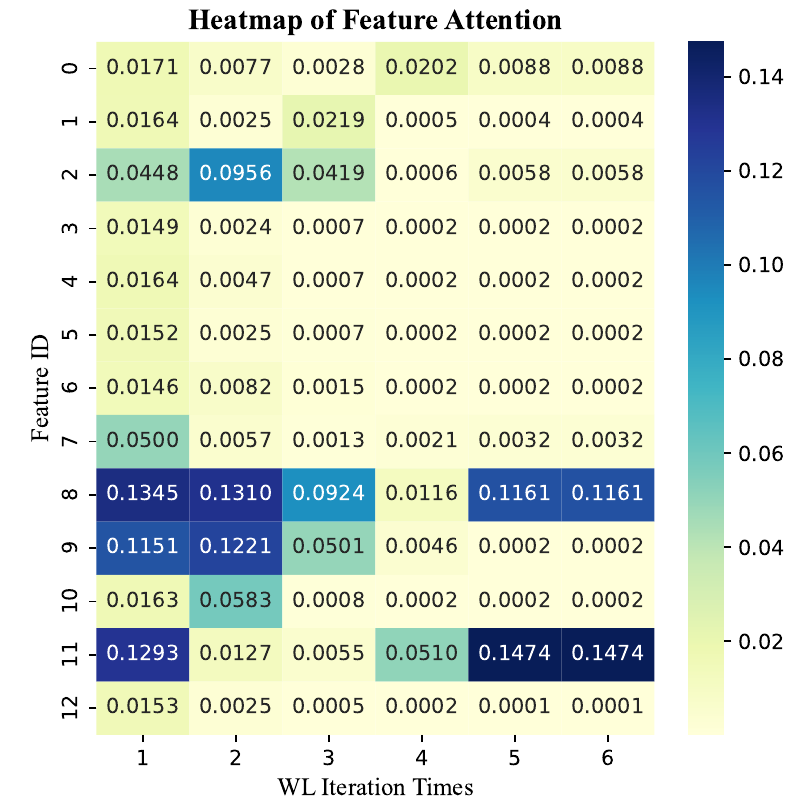}
% \vspace{-5pt}
\caption{The Attention Distribution for MUTAG.} \label{fig3}
\end{figure}
\subsubsection{\textbf{Experimental Results and Analysis}} Table ~\ref{tab4} indicates that the proposed AKBR$_i$(WL) models outperform the alternative graph deep learning methods on all five datasets. Although the accuracy of our proposed AKBR(SP) method is not the best on the MUTAG dataset, the AKBR model based on SPGK is still competitive. In fact, similar to our methods, all these alternative graph deep learning methods can also provide an end-to-end learning framework, and have more learning layers than ours. However, the proposed methods still have better classification performance than these graph deep learning methods, again demonstrating the effectiveness of the kernel-based framework, i.e., considering the common patterns shared between all sample graphs.

\subsection{The Attention Mechanism Analysis}
In this subsection, we investigate the attention distribution of different substructure invariants. For a more intuitive comparison and to further show the effectiveness of our proposed model, we visualize the attention distribution of the MUTAG dataset in Fig.~\ref{fig3}. We select the first thirteen substructure invariants, as well as feature ID 0-12. Besides, we choose the iteration times between 1 to 6. We can observe that some important substructure invariants have been assigned larger attention scores. Moreover, as the number of iterations increases, the importance of different substructures is different, demonstrating our proposed model can adaptively learn the weights of substructure invariants.

\begin{figure}
    \begin{minipage}{0.49\linewidth}
        \includegraphics[width=\textwidth]{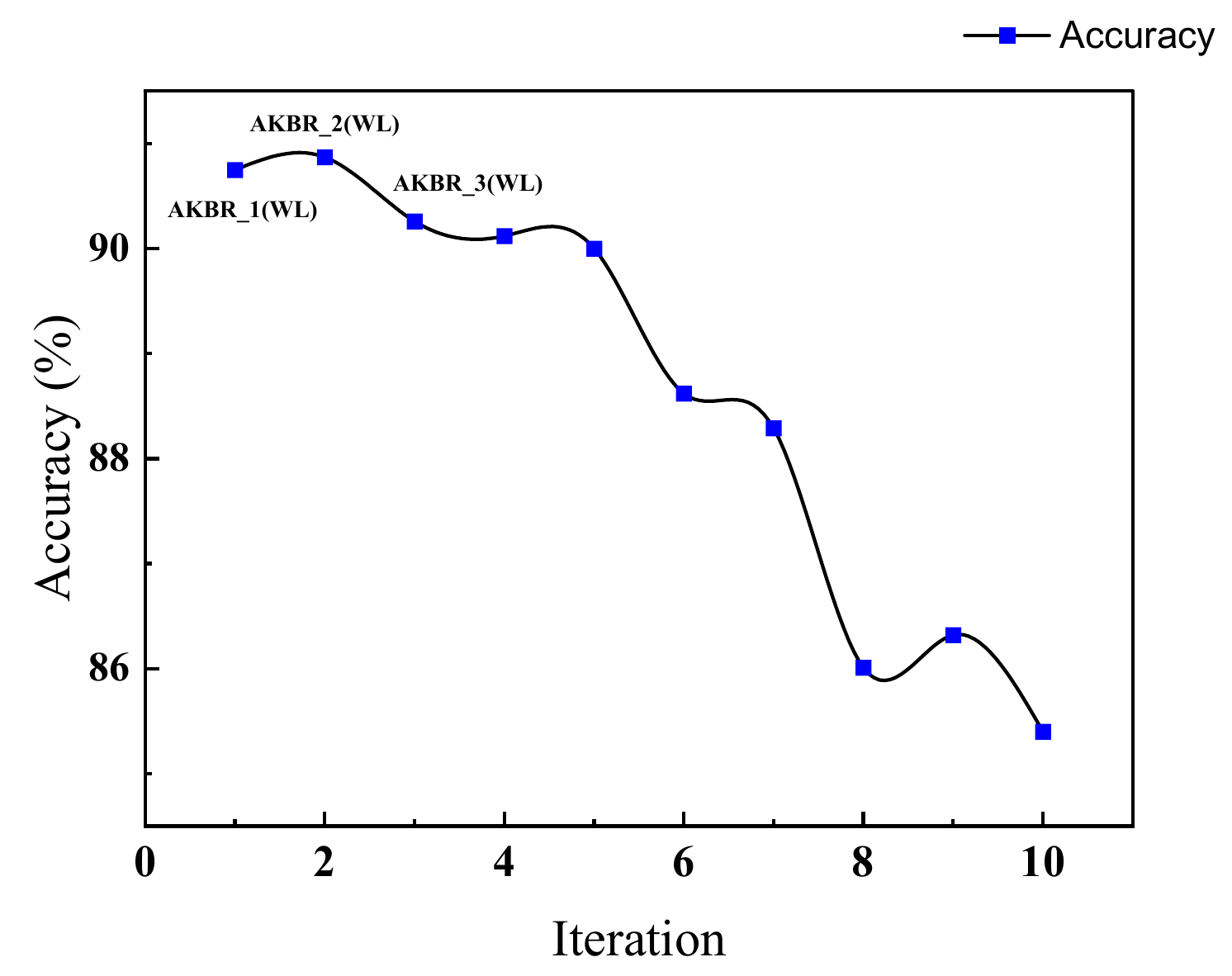}
        \centerline{(a) MUTAG}
    \end{minipage}%
    \begin{minipage}{0.49\linewidth}
        \includegraphics[width=\textwidth]{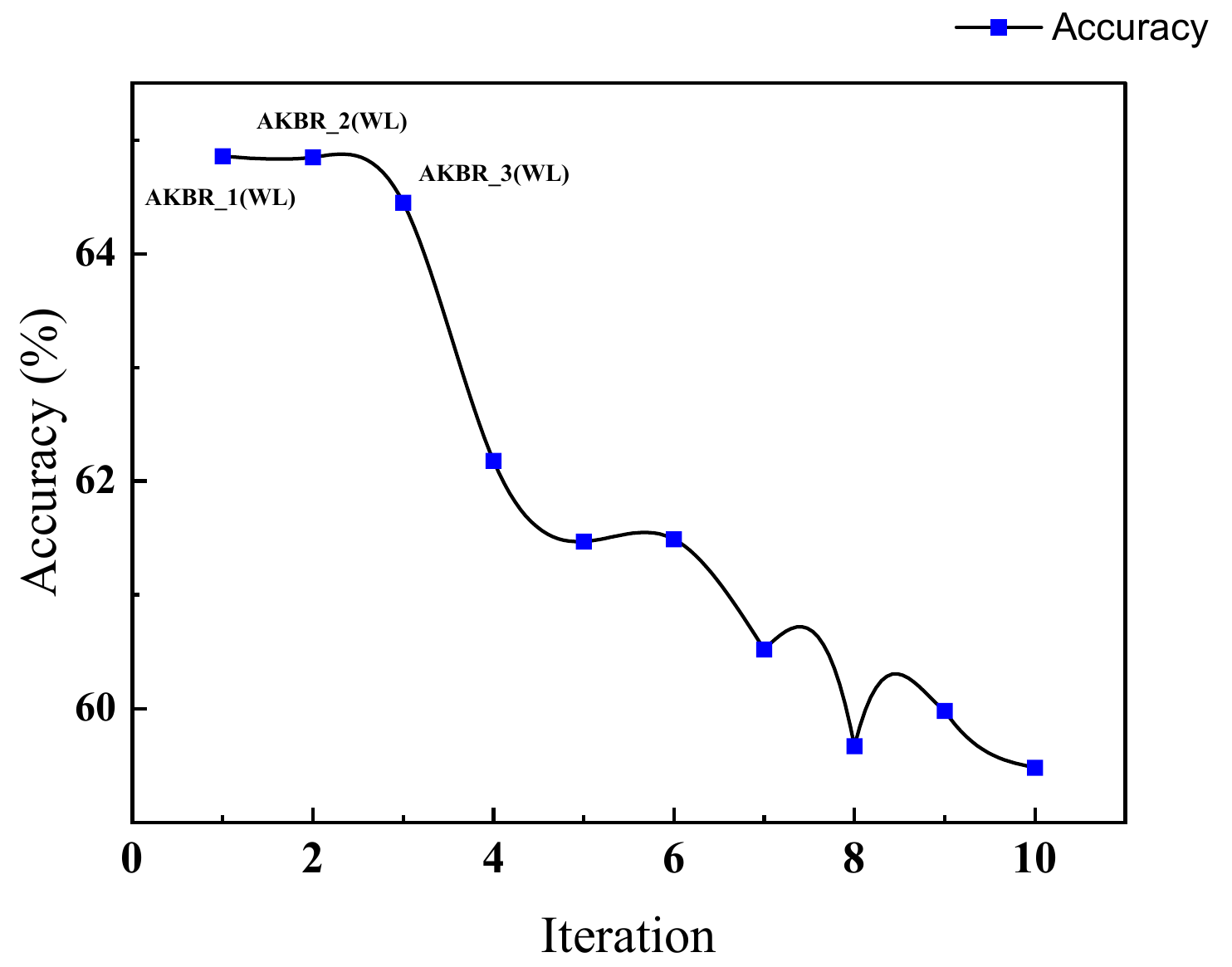}
        \centerline{(b) PTC}
    \end{minipage}
    \begin{minipage}{0.49\linewidth}
        \includegraphics[width=\textwidth]{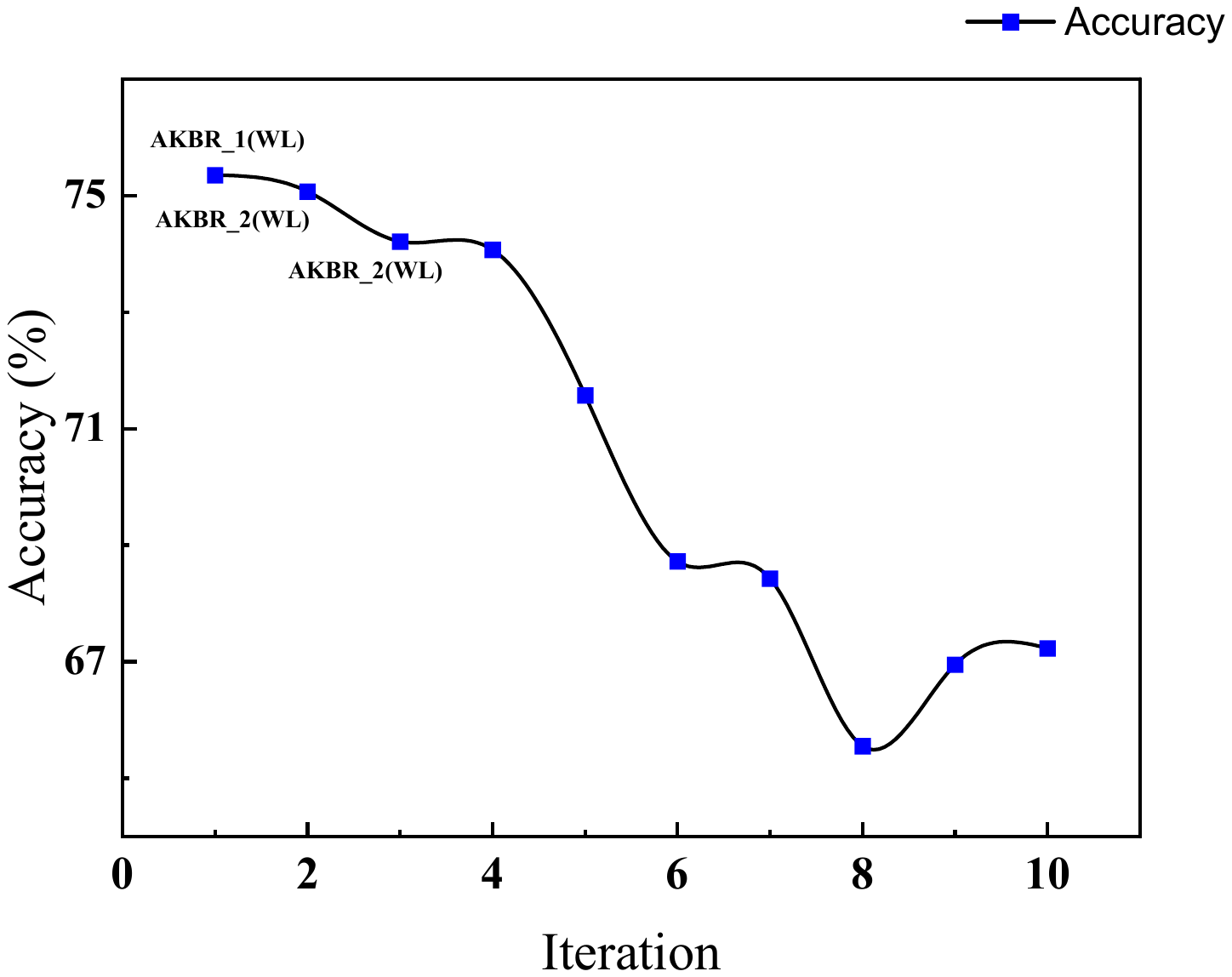}
        \centerline{(c) IMDB-BINARY}
    \end{minipage}%
    \begin{minipage}{0.49\linewidth}
        \includegraphics[width=\textwidth]{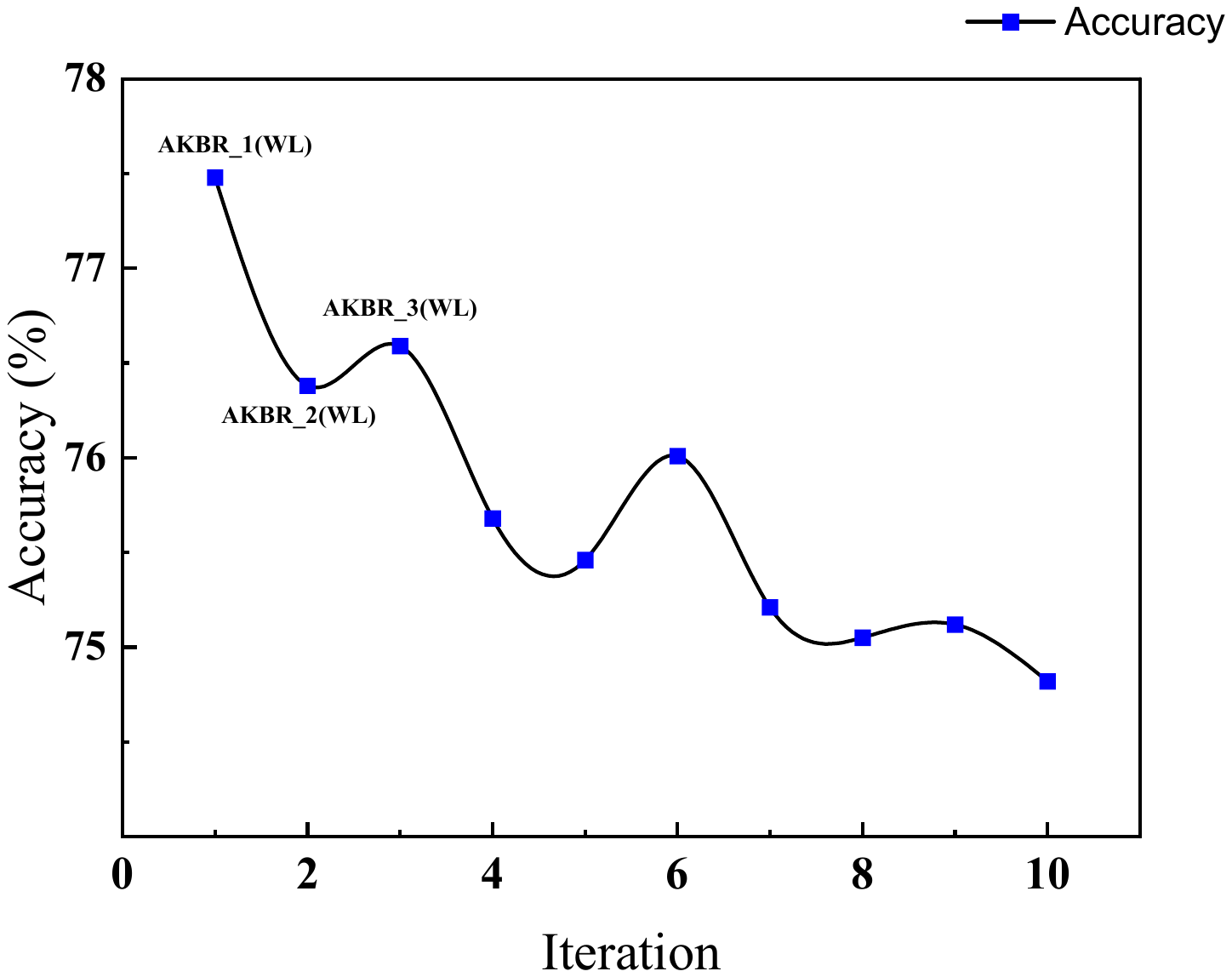}
        \centerline{(d) PROTEINS}
    \end{minipage}
        \begin{minipage}{0.49\linewidth}
        \includegraphics[width=\textwidth]{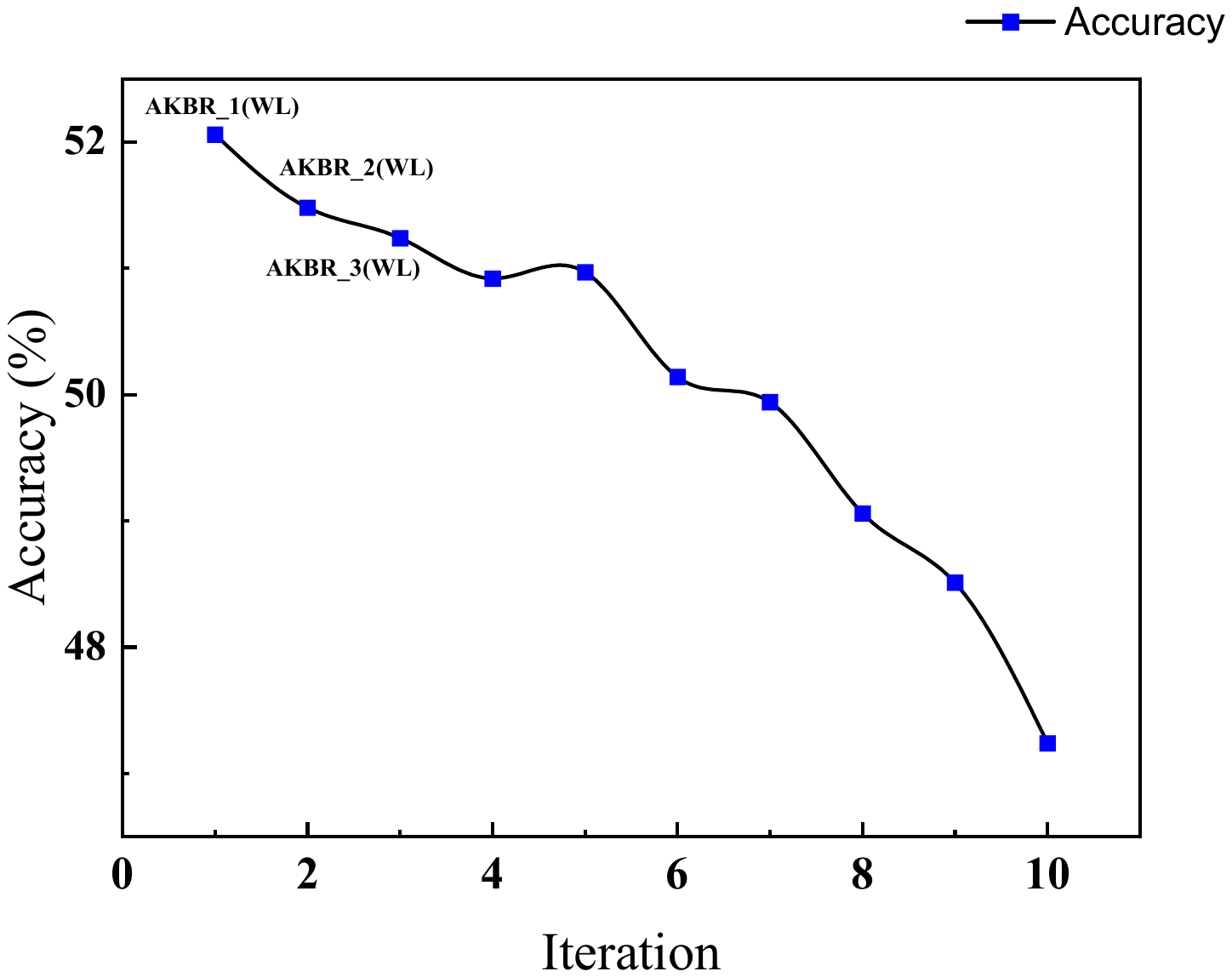}
        \centerline{(e) IMDB-MULTI}
    \end{minipage}%
    \begin{minipage}{0.49\linewidth}
        \includegraphics[width=\textwidth]{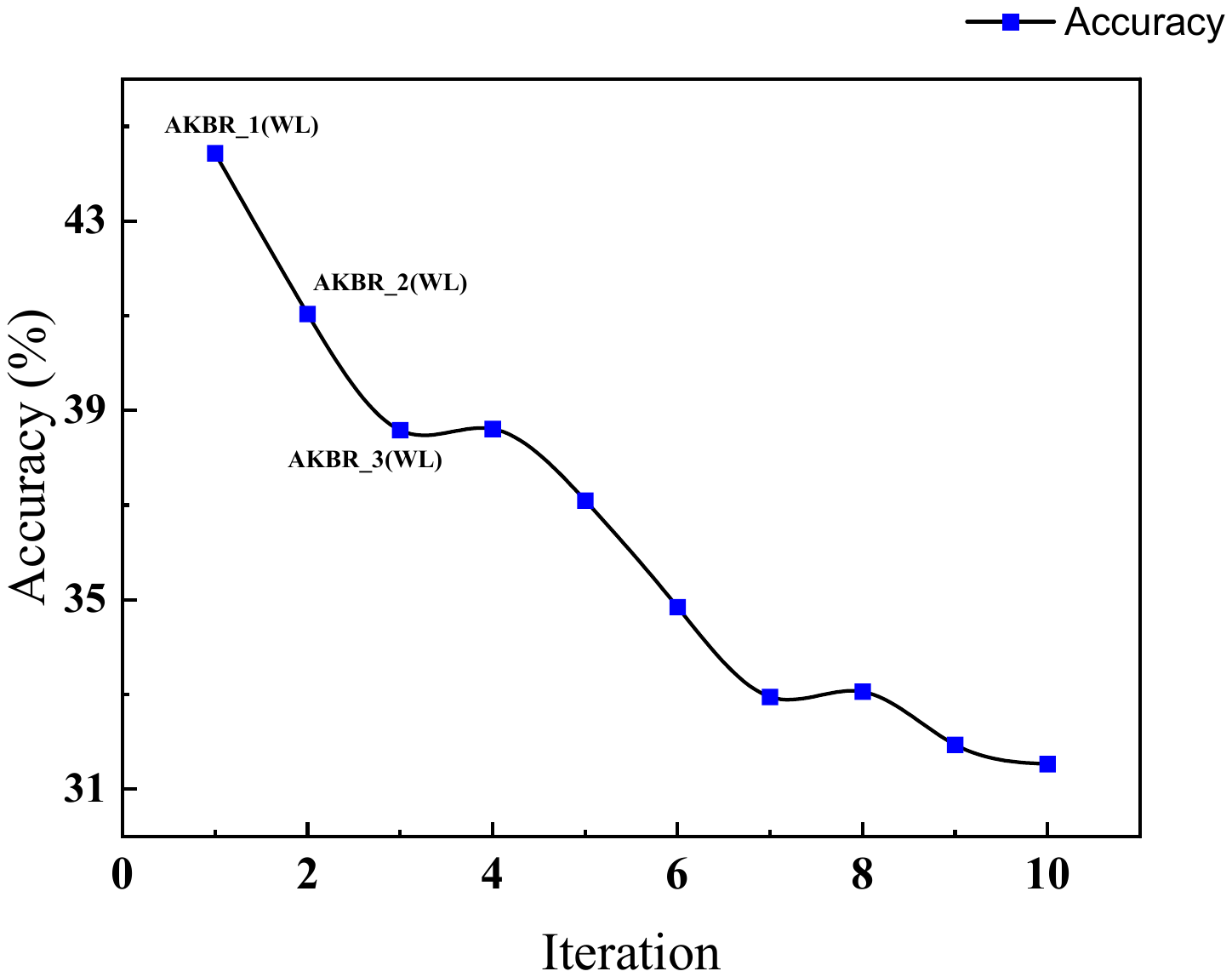}
        \centerline{(f) Shock}
    \end{minipage}
 \vspace{-5pt}
\caption{AKBR$_i$(WL) with Different Iterations} \label{fig4}
% \vspace{-15pt}
\end{figure}

\subsection{The Further Analysis for AKBR(WL)}
In this subsection, we explore how the classification accuracies of the proposed AKBR$_i$ (WL) vary with different iterations $i$ ranging from 1 to 10 on the MUTAG and PTC datasets, and the results are shown in Fig.\ref{fig4}. Note that, for other datasets we can observe the similar phenomenon. Specifically, we find that the AKBR$_i$(WL)s with the iterations $i$ from 1 to 3 usually have optimal performance. When the number of iterations further increases, the performance will be lower. On the other hand, the proposed AKBR(WL) can outperform the WLSK kernel associated with any iteration $i$. This is due to the fact that there may be more redundant feature information, when the number of substructures increases with the larger iterations. Since the proposed AKBR(WL) model can adaptively identifying the most effective substructure invariants associated with different iterations $i$, the proposed model still has better classification performance than the WLSK kernel, again demonstrating the effectiveness.

%is quite important for the classification. The observation demonstrate that the effectiveness and
% appearing in
%that are not conducive for classification. Although these substructure invariants are assigned smaller attention scores as Fig.\ref{fig3}, still generate noise for classification. Thus, the accuracy trend of classification decreases as the number of iterations increases.

%\begin{figure*}
% \centering
%%   \vspace{-10pt}
%    \subfigure{\includegraphics[width=0.28\linewidth]{MUTAG.pdf}}
%    \subfigure{\includegraphics[width=0.28\linewidth]{PTC.pdf}}
%    \subfigure{\includegraphics[width=0.32\linewidth]{IMDBB.pdf}}
%    \subfigure{\includegraphics[width=0.32\linewidth]{PROTEINS.pdf}}
%    \subfigure{\includegraphics[width=0.32\linewidth]{IMDBM.pdf}}
%    \subfigure{\includegraphics[width=0.32\linewidth]{Shock.pdf}}
%%    \vspace{-15pt}
%    \caption{AKBR$_i$(WL) with Different Iterations.}\label{fig4}
%     \vspace{-10pt}
%\end{figure*}

\section{Conclusions}
In this paper, we have proposed a novel AKBR model that can extract more effective substructure-based features and adaptively compute the kernel matrix for graph classification, through an end-to-end learning framework. Thus, the proposed AKBR model can significantly address the shortcomings arising in existing R-convolution graph kernels. Experimental results show that our proposed AKBR model outperforms the existing state-of-the-art graph kernels and graph deep learning methods. Since the proposed AKBR model can be applied to any R-convolution graph kernel, our future work is to further employ the proposed AKBR model associated with other classical graph kernels.

%\section*{Acknowledgments}
%This work is supported by the National Natural Science Foundation of China under Grants T2122020, 61976235, U21A20473, 62172370 and 61602535. Corresponding Author: Lu Bai (bailu@bnu.edu.cn; bailucs@cufe.edu.cn).

% was supervised by Dr. Lu Bai and Dr. Lixin Cui for his M.sc degree and

%-------------------------------------------------------------------------

\balance

%-------------------------------------------------------------------------
%\nocite{ex1,ex2}
%\bibliographystyle{latex12}
%\bibliography{XBib}

\bibliographystyle{IEEEtran}
\bibliography{mybib}

\end{document}